\documentclass[final]{cvpr}

\usepackage{times}
\usepackage{epsfig}
\usepackage{graphicx}
\usepackage{amsmath}
\usepackage{amssymb}
\usepackage{subfigure}

\usepackage{makecell} 

\usepackage[pagebackref=true,breaklinks=true,colorlinks,bookmarks=false]{hyperref}

\usepackage[skip=4pt]{caption}
\def\cvprPaperID{2168} 
\def\confYear{CVPR 2021}

\begin{document}

\title{Cross-MPI: Cross-scale Stereo for Image Super-Resolution \\using Multiplane Images}

\author{Yuemei Zhou\textsuperscript{1,2}, Gaochang Wu\textsuperscript{3}, Ying Fu\textsuperscript{4}, Kun Li\textsuperscript{5}, Yebin Liu\textsuperscript{1}\\
{\small \textsuperscript{1}Tsinghua University, \textsuperscript{2}ZhuoHe Tech, \textsuperscript{3}Northeastern University, \textsuperscript{4}Beijing Institute of Technology,\textsuperscript{5}Tianjin University }\\
{\tt\small zym18@mails.tsinghua.edu.cn, wugc@mail.neu.edu.com, fuying@bit.edu.cn, lik@tju.edu.cn,} \\ {\tt\small liuyebin@mail.tsinghua.edu.cn}
}

\maketitle

\begin{abstract}
Various combinations of cameras enrich computational photography, among which reference-based super-resolution (RefSR) plays a critical role in multiscale imaging systems. However, existing RefSR approaches fail to accomplish high-fidelity super-resolution under a large resolution gap, \eg, 8$\times$ upscaling, due to the lower consideration of the underlying scene structure. 
In this paper, we aim to solve the RefSR problem in actual multiscale camera systems inspired by multiplane image (MPI) representation.
Specifically, we propose Cross-MPI, an end-to-end RefSR network composed of a novel plane-aware attention-based MPI mechanism, a multiscale guided upsampling module as well as a super-resolution (SR) synthesis and fusion module.
Instead of using a direct and exhaustive matching between the cross-scale stereo, the proposed plane-aware attention mechanism fully utilizes the concealed scene structure for efficient attention-based correspondence searching. Further combined with a gentle coarse-to-fine guided upsampling strategy, the proposed Cross-MPI can achieve a robust and accurate detail transmission.
Experimental results on both digitally synthesized and optical zoom cross-scale data show that the Cross-MPI framework can achieve superior performance against the existing RefSR methods and is a real fit for actual multiscale camera systems even with large-scale differences.
\end{abstract}

\begin{figure}[t]
\begin{center}
\includegraphics[width=1.0\linewidth]{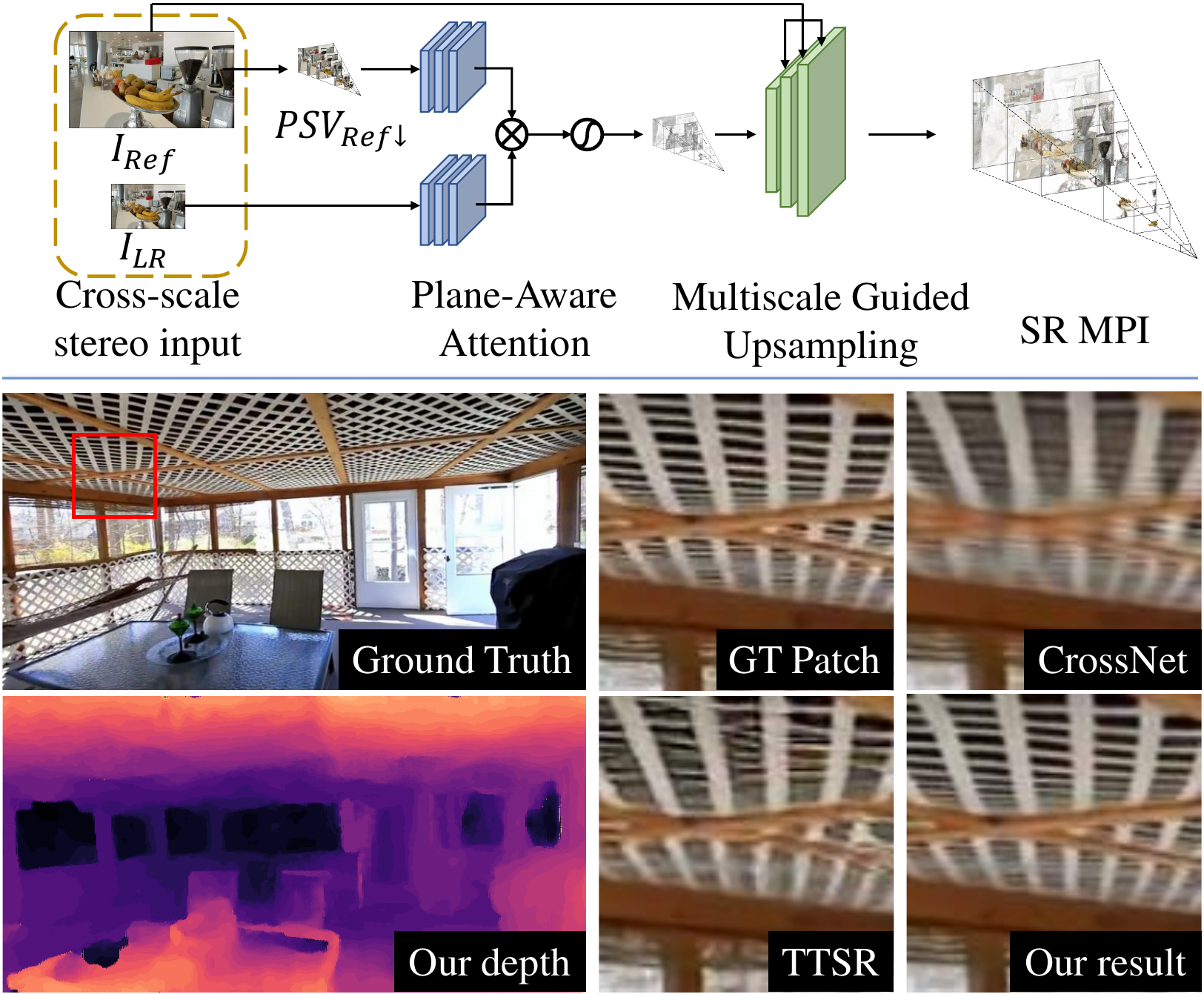}
\end{center}
\vspace{-4mm}
\caption{\textbf{Top}: Our Cross-MPI network takes cross-scale stereo images as input and generates super-resolved MPIs through a delicate plane-aware attention mechanism and an effective multiscale guided upsampling module. \textbf{Bottom}: An 8$\times$ SR comparison between Cross-MPI and the state-of-the-art RefSR work, CrossNet~\cite{zheng2018crossnet} and TTSR~\cite{yang2020learning}, along with our intermediate depth estimation result.}
\label{fig:teaser}
\vspace{-1mm}
\end{figure}

\section{Introduction}

With computational photography prospering over the past few decades, higher-quality and fidelity imaging is in sustained demand for both industry and daily life. Hybrid multiscale imaging camera systems~\cite{wang2017the, 2017Multiscale, iphone11pro, HUAWEIP40proplus} stand out for their capability to compose different camera characteristics and obtain a much higher spatial bandwidth. 
Hybrid multiscale imaging has been successfully applied in various imaging systems, including hybrid light field cameras~\cite{wang2017the}, gigapixel video systems~\cite{2017Multiscale}, and mobile phones with multi-camera systems~\cite{iphone11pro, HUAWEIP40proplus}. 
To realize hybrid multiscale imaging, one of the key issues is how to match and fuse cross-resolution images from different cameras under certain perspective parallax.

Reference-based image super-resolution (RefSR) has been proposed to solve the cross-scale reconstruction problem and applied to gigapixel video systems~\cite{2017Multiscale} or hybrid light field imaging systems~\cite{boominathan2014improving, wang2017the}. 
RefSR aims to super-resolve a low-resolution (LR) image through additional high-resolution (HR) reference image(s).
Unlike data-driven single image super-resolution (SISR)~\cite{lim2017enhanced,zhang2018residual,zhang2018image,ledig2017photo,ma2020structure}, which only utilizes prior knowledge learned from massive data to build an upsampling model, RefSR can produce much more satisfactory results due to the additional HR references, especially for upsampling factors larger than 4$\times$. 

Currently, most RefSR methods typically follow a ``warp-and-synthesis'' pipeline, which can be further divided into two branches based on different detail transfer schemes, including pixel-alignment and patch-matching RefSR. For the pixel-alignment RefSR, the warping process is conducted based on a certain pixel-to-pixel mapping between two images, \eg, optical flow~\cite{zheng2018crossnet} and depth map~\cite{wang2017the, zhao2018cross}. Nevertheless, pixel-alignment RefSR would fail when dealing with large-parallax cases or image pairs under a large resolution gap due to the lack of ability to capture long-distance correspondences. For the patch-matching RefSR~\cite{boominathan2014improving, 2017Learning, wang2017the, zhang2019image, yang2020learning}, image patches are extracted to locate correspondences and blended accordingly for HR detail transmission. However, it is limited by complex textures or textureless areas and inherently generates grid artifacts due to averaging window blurs. 
To summarize, existing RefSR methods neglect the underlying scene structure of cross-scale views, and cannot capture correspondences and transfer high-fidelity details with large resolution gaps (\eg, 8$\times$) and large parallax.


In this paper, we set sights on an emerging scene representation and view synthesis framework, \ie, multiplane images (MPIs), to solve the cross-scale RefSR problem in actual multiscale camera systems. Thanks to the MPIs, the performance of image-based novel view synthesis~\cite{zhou2018stereo, tucker2020single, srinivasan2019pushing, duvall2019compositing, mildenhall2019local} has drastically improved in the last few years.
The MPI representation has many good properties, including the ability to capture concealed surfaces from the input views and the adaptability to discontinuous edges and transparent scenes. 
However, the MPI estimation requires image pairs to have the same resolution with slight viewpoint changes, which is not available in RefSR tasks, especially when the resolution gap of the input pair is up to 8$\times$.

To fully understand the scene structure, we propose an end-to-end MPI-based approach that fully explores the properties of cross-scale image pairs by leveraging vanilla MPIs in an efficient multiscale guided coarse-to-fine pipeline. First, we introduce a novel MPI representation established on a lightweight plane-aware attention module. Compared with the original MPI estimation network with simple concatenation and convolution, our plane-aware attention mechanism explicitly captures correspondences along depth planes. Moreover, compared with popular non-local attention mechanisms~\cite{wang2018non, fu2019dual, wang2019learning, yang2020learning}, which have to maintain a large matrix for exhaustive pixel-to-pixel comparison, it takes much fewer computations and less storage space. Second, to estimate robust and detailed correspondences in a space with a considerable resolution difference, \eg, 8$\times$, we design a multiscale guided upsampling module. This module takes the correspondences captured in LR space as input and conducts gentle correspondence upsampling with guidance from the HR reference view in a coarse-to-fine manner, producing super-resolved MPIs.
Finally, accurate transmission and fusion of high-quality details can be accomplished utilizing the obtained super-resolved MPI representation. 
The entire pipeline is designed in an end-to-end fashion and can infer high-fidelity depth information in an unsupervised manner. Experimental results show that our method is visually and quantitatively superior to existing methods with high computational efficiency.
We also build a real hybrid cross-scale stereo system and collect zooming data to validate the superior performance of the proposed algorithm. The main contributions of this paper are as follows:

\begin{itemize}
\setlength{\itemsep}{0pt}
\setlength{\parsep}{0pt}
\setlength{\parskip}{0pt}
\vspace{-0.2cm}
\item We take a close look at the RefSR problem through the lens of the MPI representation, achieving super-resolution up to 8$\times$ on real hybrid imaging systems.
\item We propose a novel plane-aware attention mechanism for MPI estimation that can achieve more explicit and efficient correspondence estimation compared with the original direct concatenation-and-convolution operation.

\item We propose a novel multiscale guided upsampling module for cross-scale multiplane image synthesis that can solve the
matching problem under large resolution differences. A fine-detailed depth map that encodes the scene structure can also be inferred.
\end{itemize}

\section{Related Work}
\label{section:re}

In this section we review previous work on single image super-resolution (SISR) and reference-based image super-resolution (RefSR), which are the most relevant to our work. 

\subsection{Single Image Super-Resolution}

In recent years, learning-based methods have dominated the SISR research community due to the pioneering work of SRCNN~\cite{dong2014learning} proposed by Dong \etal, which directly maps LR images to HR ground truth by a 3-layer CNN.
Kim \etal~\cite{kim2016accurate, kim2016deeply} further improved the performance by equipping deeper networks. 
A series of works including SRResNet~\cite{ledig2017photo}, EDSR~\cite{lim2017enhanced} and RDN~\cite{zhang2018residual}, leverage residual learning~\cite{he2016deep} to boost the development of SISR.
Zhang \etal~\cite{zhang2018image} proposed improving residual learning by a channel attention mechanism and built a deep model called RCAN to further improve the performance. 
However, these methods simply focus on raw pixel comparison, \eg, mean square error (MSE) for model optimization, which tends to produce blurry images. To mitigate this problem, Johnson \etal~\cite{johnson2016perceptual} first introduced the high-level perceptual loss into SR tasks to mimic the human perception system.
After that, SRGAN~\cite{ledig2017photo} adopted generative adversarial networks (GANs)~\cite{goodfellow2014generative} to generate photorealistic SR results. Sajjadi \etal~\cite{sajjadi2017enhancenet} proposed EnhanceNet, which utilizes gram matrix-based texture matching loss to enhance local similar textures. Furthermore, ESRGAN~\cite{wang2018esrgan} enhanced SRGAN by introducing a residual-in-residual dense block (RRDB) and relativistic adversarial loss. Moreover, RSRGAN~\cite{zhang2019ranksrgan} improved SR visual quality by a ranker that can learn the behaviour of perceptual metrics and a novel rank-content loss to optimize the perceptual quality. More recently, Ma \etal~\cite{ma2020structure} proposed SPSR, which exploits gradient priors and structural information of images to achieve better visual quality.


\subsection{Referenced-based Image Super-Resolution}

Compared with SISR, which directly predicts missing HR details from the LR input, RefSR benefits from additional referable HR inputs. Since the decisive factor for RefSR to achieve good performance is the accuracy of matching and fusion between the LR input and HR reference, we can classify RefSR work into pixel-alignment-related and patch-matching-related. For the pixel-alignment RefSR, pixel-to-pixel matching and warping are estimated and operated for HR reference registration. Early work, such as Landmark~\cite{yue2013landmark}, was proposed to retrieve correlated web images as a reference, and a global registration and an energy minimizing problem were solved to recover the desired SR image. Multiscale gigapixel video~\cite{2017Multiscale} proposes achieving cross-scale view alignment in the gigapixel video system through mesh-based warping and optical flow warping. However, mesh-based warping limits the deformation when large parallax occurs. Except for warping based on matched feature points, the work of reconstructing light fields with extra HR references, such as~\cite{wang2017the, zhao2018cross}, further introduces depth map estimation for transferring finer high-frequency details. However, these methods focus on light field depth estimation and cannot be used to robustly find cross-scale correspondences for unstructured camera arrays such as~\cite{2017Multiscale}. Recently, CrossNet~\cite{zheng2018crossnet} proposed a learning-based method to conduct cross-scale warping by predicting optical flow, which achieves high accuracy on light field datasets but fails when dealing with large-parallax cases.
These pixel-alignment RefSR methods heavily depend on the pixel-level alignment quality between the LR and HR references, where optical flow estimation and feature point matching are limited to small-parallax conditions.

Furthermore, some works~\cite{boominathan2014improving, 2017Learning, wang2017the} utilize image patches to locate correspondences and synthesize HR details from references. Despite the favourable performance achieved by these methods, they are limited to the large search space of patch-based methods and tend to induce the problem of blending blur as a result of averaging the transferred HR patches. 
Further expanding references to HR images without view constraints such as light field applications, SRNTT~\cite{zhang2019image} and TTSR~\cite{yang2020learning} combine patch matching with advanced deep learning architectures to achieve visually pleasant SR results. The former work reconsiders RefSR as a texture transfer problem and synthesizes swapped pre-trained VGG features in a style transfer way, while the latter designs a texture transformer fully utilizing an attention mechanism to accomplish matching and synthesis. Both works require exhausting comparisons between 
the extracted patches and cannot handle image areas with fewer or repeating textures. 


\begin{figure}[t]
\begin{center}
\includegraphics[width=0.9\linewidth]{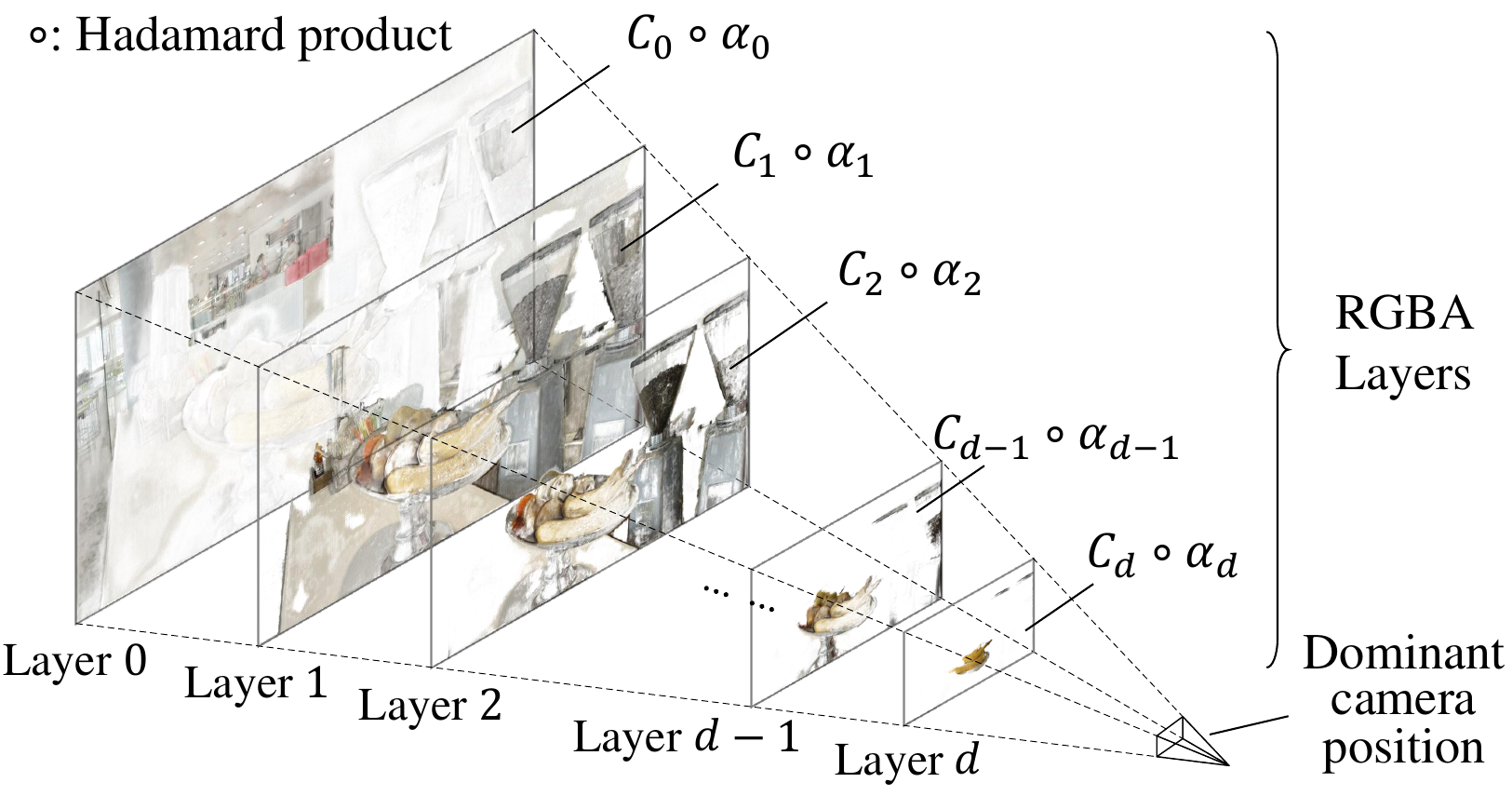}
\end{center}
\vspace{-4mm}
\caption{An illustration of MPI representation.}
\label{mpi}
\end{figure}



\begin{figure*}
\begin{center}
\includegraphics[width=0.98\linewidth]{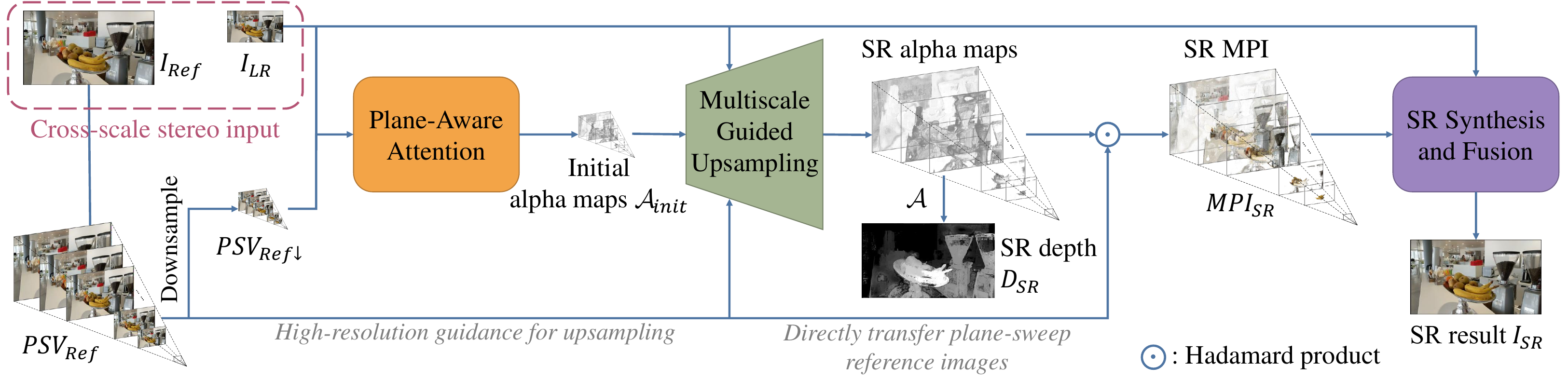}
\end{center}
\vspace{-3mm}
\caption{The Cross-MPI pipeline. We first estimate the initial alpha maps through a plane-aware attention-based MPI module. A novel multiscale guided upsampling module is then designed for generating super-resolved alpha maps. To transfer HR details of the reference view and generate the final SR result, the pipeline ends with an SR synthesis and fusion module. 
The whole pipeline is elaborately designed for the cross-scale stereo RefSR problem by fully considering camera relationships as well as the underlying scene structure.}
\label{fig:pipeline}
\vspace{-3mm}
\end{figure*}


\section{Approach}
\label{section:app}

Given a cross-scale stereo camera system, our goal is to use the HR reference image $I_{Ref}$ to super-resolve the corresponding LR image $I_{LR}$. We formulate the training tuple as $\langle I_{LR}, I_{Ref}, c_{LR}, c_{Ref}, I_{GT} \rangle$, where $I_{LR}\in\mathbb{R}^{h\times{w}\times{c}}$, $I_{Ref}\in\mathbb{R}^{\beta h\times{\beta w}\times{c}}$, $\beta$ is the resolution difference factor, $c_{LR}$ and $c_{Ref}$ are calibration parameters (including intrinsics and extrinsics) of the input cross-scale stereo pair, and $I_{GT}\in\mathbb{R}^{\beta h\times{\beta w}\times{c}}$ is the super-resolution ground truth of the LR view. To decode the captured scene components into planes, we also calculate 
a plane sweep volume $PSV_{Ref} = \{PSI_{Ref}^i\}, i = 1, ..., d$ from $\{I_{Ref}, c_{LR}, c_{Ref}\}$, where $d$ is the number of depth planes.

Different from traditional MPI estimation, our stereo input suffers from a large resolution difference, 8$\times$ for instance. To match the frequency band of $I_{LR}$ and obtain alpha maps with target resolution and fine details, we first calculate rough alpha maps at low resolution using plane-aware attention as depicted in Sect.~\ref{section:plane-aware} and gradually add details through the proposed multiscale guided upsampling module described in Sect.~\ref{section:csmpi}. Then, the following SR synthesis and fusion module depicted in Sect.~\ref{section:fusion} is used to adequately transfer details from the HR reference and generate the final SR result. In addition, loss functions for optimizing the proposed Cross-MPI network and the implementation details are discussed in Sect.~\ref{section:loss} and Sect.~\ref{section:implementation}. Please see Fig.~\ref{fig:pipeline} for an overview.


\subsection{Plane-Aware Attention-based MPIs}
\label{section:plane-aware}

Here, we introduce our innovative plane-aware attention mechanism for efficient feature matching. The scene representation we adopt, multiplane images (MPIs)~\cite{zhou2018stereo}, is composed of multiple fronto-parallel planes at a fixed depth range relative to a selected view. The MPI representation is a collection of RGBA layers $\{(C_1, \alpha_1), ..., (C_d, \alpha_d)\}$, where $C$ indicates RGB colour, $\alpha$ indicates occupancy of each plane and $d$ is the number of depth planes. Please see Fig.~\ref{mpi} for more details. As can be know from this representation, the set of alpha maps, $\mathcal{\mathcal{A}}=\{\alpha_i\}, i = 1, ..., d$, indicates the occupancy of colours on each depth plane and reflects the softness of the decoded scene layers, especially for boundaries and reflective or transparent parts. In our RefSR task, the alpha maps are perfect for transferring HR details from the reference view to the LR view.

\begin{figure}[t]
\begin{center}
\includegraphics[width=1.0\linewidth]{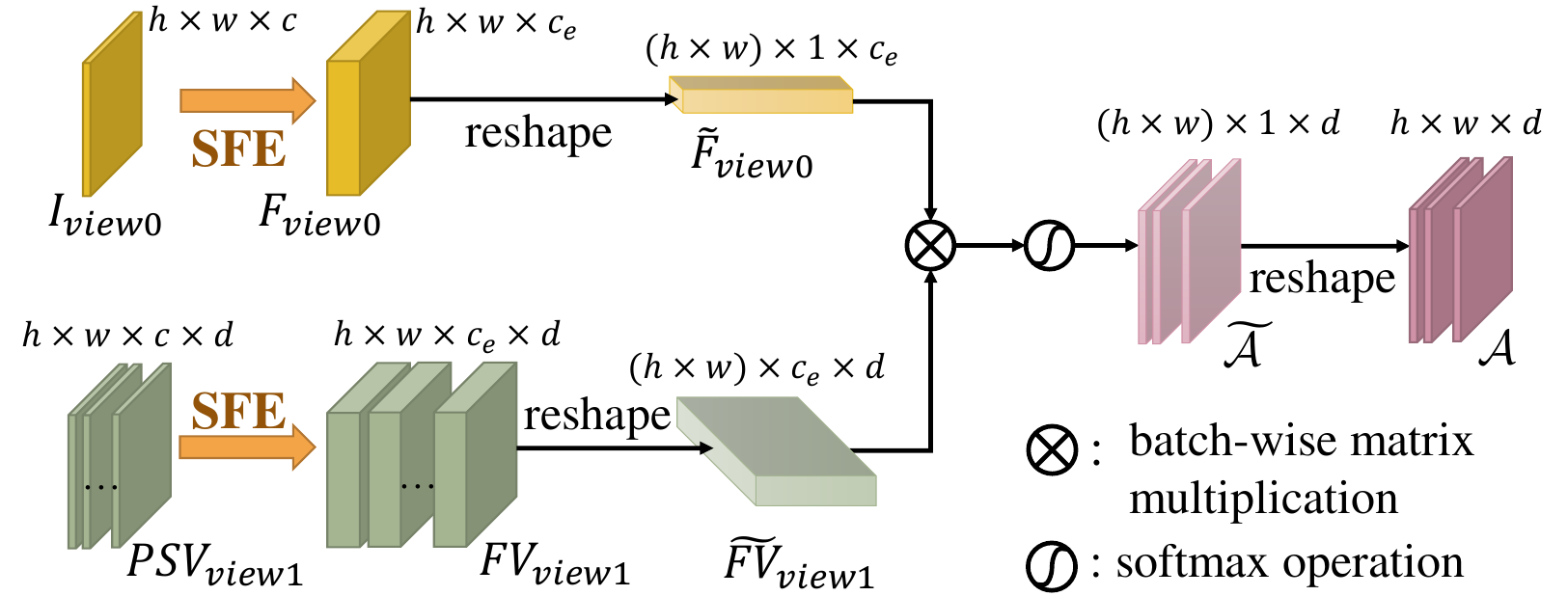}
\end{center}
\vspace{-4mm}
\caption{Our proposed novel plane-aware attention mechanism explicitly calculates the correlation of the input view and the corresponding plane sweep image on each image plane.}
\label{fig:plane-ware}
\end{figure}

However, MPIs are normally estimated by a CNN whose input is the concatenation of one dominant view and the plane sweep volume of the other view. 
Other than learning alpha maps simply through the stack of network layers with implicit and weak comparisons, we develop a plane-aware attention-based MPI mechanism to achieve explicit matching, as depicted in Fig.~\ref{fig:plane-ware}. Inspired by the strong capability of attention to capture long-range dependencies in computer vision tasks, such as image segmentation~\cite{fu2019dual}, stereoSR~\cite{wang2019learning} and RefSR~\cite{yang2020learning}, our plane-aware attention further considers the inherent principle of the MPI estimation procedure from stereo input and explicitly conducts an adequate and efficient comparison on each depth plane to estimate multiplane correspondences.

\noindent {\bf Shared Feature Extractor.} 
We first extract dense features from the LR inputs $(I_{LR}, PSV_{Ref\downarrow})$ through a shared feature extractor (SFE), which is a residual atrous spatial pyramid pooling (ResASPP) module~\cite{wang2019learning}. Given $I_{LR}\in\mathbb{R}^{h\times{w}\times{c}}$ and $PSV_{Ref\downarrow}\in\mathbb{R}^{h\times{w}\times{c}\times{d}}= \{PSI_{Ref\downarrow}^i\}, i = 1, ..., d$ where $PSI_{Ref\downarrow}^i$ is calculated by $1/\beta \times$ bicubic downsampling to match the LR domain, the feature extraction process can be expressed as:
\begin{equation}
\begin{aligned}
F_{LR} &= \mathrm{SFE}(I_{LR}),  \\
FV_{Ref\downarrow} &= \mathrm{Concat}(\{\mathrm{SFE}(PSI_{Ref\downarrow}^i)\}_{i=1}^d),  \label{con:feature_extractor}
\end{aligned}
\end{equation}
where $F_{LR}\in\mathbb{R}^{h\times{w}\times{c_e}}$, $FV_{Ref\downarrow}\in\mathbb{R}^{h\times{w}\times{c_e}\times{d}}$, $c_e$ is the channel size of the feature maps, $\mathrm{SFE}(\cdot)$ is the shared feature extractor and $\mathrm{Concat}(\cdot)$ is the channel-wise concatenation. In addition, the receptive field of this feature extractor is large enough for further plane-aware attention formation based on the structure of the ResASPP module.


\noindent {\bf Plane-Aware Attention.} Once the rich features of the LR images to be matched are extracted, novel plane-aware attention is proposed to accomplish precise matching. 
As illustrated in Fig.~\ref{fig:plane-ware}, given the features at the LR view $F_{LR}$ and plane sweep features of the reference view $FV_{Ref\downarrow}$, we reshape them into $\widetilde{F}_{LR}\in\mathbb{R}^{n\times{1}\times{c_e}}$ and $\widetilde{FV}_{Ref\downarrow}\in\mathbb{R}^{n\times{c_e}\times{d}}$, where $n=h\times{w}$ is the number of pixels. After that, we perform a batch-wise matrix multiplication between $\widetilde{F}_{LR}$ and $\widetilde{FV}_{Ref\downarrow}$ and apply a softmax layer to obtain the plane-aware attention $\widetilde{\mathcal{A}}_{init}\in\mathbb{R}^{n\times{1}\times{d}}$, which is then reshaped to be the initial estimation of the alpha maps $\mathcal{A}_{init}\in\mathbb{R}^{h\times{w}\times{d}}$. The plane-aware attention estimation procedure can be expressed as:
\begin{equation}
\begin{aligned}
\widetilde{\mathcal{A}}_{init} &= \mathrm{softmax}(\widetilde{F}_{LR}\otimes\widetilde{FV}_{Ref\downarrow}), \\
\mathcal{A}_{init} &= \mathrm{reshape}(\widetilde{\mathcal{A}}_{init}),  \label{con:plane-aware_attention}
\end{aligned}
\end{equation}
where $\otimes$ denotes batch-wise matrix multiplication. The initial alpha maps $\mathcal{A}_{init}$ represent the similarities between $I_{LR}$ and $PSI_{Ref\downarrow}^i$ on each depth plane. The proposed plane-aware attention effectively decomposes scene components into predefined planes.


\subsection{Multiscale Guided Upsampling Module}
\label{section:csmpi}

\begin{figure}[t]
\begin{center}
\includegraphics[width=1.0\linewidth]{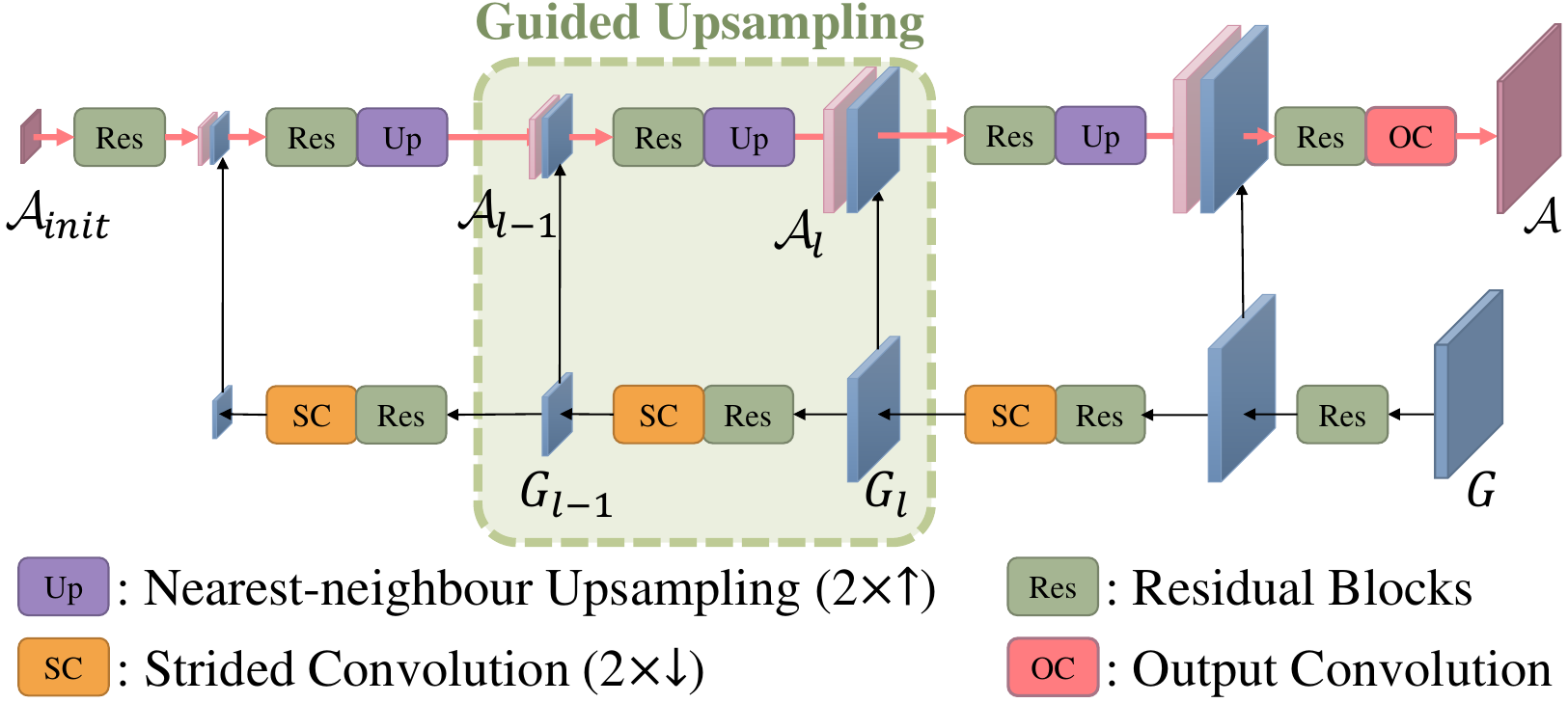}
\end{center}
\vspace{-4mm}
\caption{The multiscale guided upsampling module gradually enriches details of the initial alpha maps by introducing multiscale guidance from the reference view. For the operation of each level as depicted in the dashed green box, please refer to Eq.~(\ref{con:decoder}).} 
\label{fig:guided}
\end{figure}

Benefiting from plane-aware attention-based MPIs in LR space, our method can efficiently obtain the initial alpha maps. To further upsample the alpha maps into the target spatial size, we design a multiscale guided upsampling module, which can add more details in a coarse-to-fine manner for more precise mapping. We concatenate $I_{LR\uparrow}$ and $PSV_{Ref}$ as guidance while gradually upsampling the initial alpha maps $\mathcal{A}_{init}\in\mathbb{R}^{h\times{w}\times{d}}$, where $I_{LR\uparrow}$ is calculated by $\beta\times$ bicubic upsampling of $I_{LR}$ to match the HR domain. As illustrated in Fig.~\ref{fig:guided}, we use a multiscale guided filter within the decoder of $\mathcal{A}_{init}$, and features are progressively fused into the up-sampling procedure. The multiscale guided upsampling module can be represented as:
\begin{equation}
\begin{aligned}
G_{l-1} &= [\mathrm{Res}(G_l)]\downarrow_{2\times},  \\
\mathcal{A}_l &= [\mathrm{Res}(\mathrm{Concat}(\mathcal{A}_{l-1},G_{l-1}))]\uparrow_{2\times},  \label{con:decoder}
\end{aligned}
\end{equation}
where $G_l$ is the guidance map at spatial level $l$ with the maximum level guidance map indicating $\mathrm{Concat}(I_{LR\uparrow}, PSV_{Ref})$, $\mathcal{A}_l$ is the attention feature map at level $l$ with the minimum level feature map as input $\mathcal{A}_{init}$ and $\mathrm{Res}(\cdot)$ denotes residual blocks. Notation $\mathrm{Concat}(\cdot)$ denotes channel-wise concatenation, and $\downarrow_{2\times}$ and $\uparrow_{2\times}$ denote the strided convolution and nearest-neighbour upsampling, respectively. 
In the output layer of the multiscale guided upsampling module, a $1\times1$ convolution is applied to recover the channel size of the final alpha maps as:
\begin{equation}
\mathcal{A} = \mathrm{softmax}(\mathrm{Conv}_{1\times1}(\mathrm{Res}(\mathrm{Concat}(\mathcal{A}_{l}, G_{l})))),  \label{con:decoder_out}
\end{equation}
where the alpha maps $\mathcal{A}$ have a spatial size of $\beta h\times{\beta w}\times{d}$, and the softmax operation ensures the depth probabilities sum to one on the emitting light of each pixel unit.

\subsection{SR Synthesis and Fusion}
\label{section:fusion}
After we obtain the alpha maps $\mathcal{A}$ with the same SR spatial size, we can generate SR MPIs of the LR view. The SR MPIs only contain HR colours from the HR reference view, \ie, HR plane sweep images element-wise multiplying the alpha maps on each depth plane. This procedure can be presented as:
\begin{equation}
MPI_{SR} = PSV_{Ref} \circ \mathcal{A}, \label{con:sr-mpi}
\end{equation}
where $MPI_{SR}\in \mathbb{R}^{h\times{w}\times{c}\times{d}}$, and $\circ$ denotes the Hadamard product. 

By element-wise summarizing $MPI_{SR}$ through the depth plane channel, a transferred HR image $T_{Ref}\in \mathbb{R}^{h\times{w}\times{c}}$ at the LR view can be synthesized. To further fuse the synthesized fine details according to the LR input and produce a final SR result, we then design an SR fusion sub-network. The SR synthesis and fusion process can be expressed as:
\begin{equation}
\begin{aligned}
T_{Ref} &= \sum_i MPI_{SR}^i, \\
I_{SR} &= \mathrm{FuseNet}(T_{Ref}, I_{LR\uparrow}),  \label{con:sythesis_and_fuse}
\end{aligned}
\vspace{-1mm}
\end{equation}
where $MPI_{SR}^i\in \mathbb{R}^{h\times{w}\times{c}}$ is the $i-$th component on the depth plane channel, $I_{LR\uparrow}$ is calculated by $\beta\times$ bicubic upsampling, and $\mathrm{FuseNet}(\cdot)$ denotes the SR fusion sub-network, which is a residual network with multiple cascaded sub-residual blocks. Above all, the SR synthesis and fusion module are designed to adequately transfer HR textures to align with the LR input. 

The whole pipeline is trained in an end-to-end way, and we can additionally obtain the scene depth $D_{SR}\in\mathbb{R}^{\beta h\times\beta w\times{1}}$ from the alpha maps by simply applying an argmax function:
\begin{equation}
D_{SR} = \mathop{\arg \max}\limits_{i}(\mathcal{A}^{\beta h\times\beta w\times i}). \label{con:depth}
\vspace{-2mm}
\end{equation}

We also provide some visual results of our super-resolved depths in Fig.~\ref{fig:ablation}.

\subsection{Loss Functions}
\label{section:loss}

The design of our overall loss considers maintaining the spatial characteristics of the SR image, gaining better visual quality and supervising within the multiscale structure to obtain precise alpha maps. Specifically, the loss is composed of the following three parts: reconstruction loss, perception loss and internal supervision loss:
\begin{equation}
\mathcal{L}_{all} = \lambda_{rec}\mathcal{L}_{rec} + \lambda_{per}\mathcal{L}_{per} + \lambda_{is}\mathcal{L}_{is}.  \label{con:loss_all}
\end{equation}

\begin{table*}
\begin{center}
\caption{Quantitative evaluation (PSNR/SSIM) on the RealEstate10K dataset~\cite{zhou2018stereo} under $8\times$ resolution gap.}
\label{table:realestate_num}
\begin{tabular}{l|c c c c c}
\hline
Algorithm & frame diff. = 9 & frame diff. = 11 & frame diff. = 15 & frame diff. = 19 & frame diff. = 23\\
\hline
Bicubic & 25.016 / 0.732 & 25.017 / 0.733 & 25.052 / 0.735 & 25.044 / 0.734 & 25.052 / 0.734\\
MDSR~\cite{mdsr} & 27.424 / 0.808 & 27.434 / 0.808 & 27.490 / 0.810 & 27.480 / 0.810 & 27.486 / 0.810\\
RCAN~\cite{zhang2018image} & 27.846 / 0.822 & 27.855 / 0.822 & 27.918 / 0.824 & 27.907 / 0.824 & 27.908 / 0.824\\
ESRGAN~\cite{wang2018esrgan} & 25.555 / 0.743 & 25.563 / 0.743 & 25.633 / 0.746 & 25.634 / 0.746 & 25.633 / 0.746\\
RankSRGAN~\cite{zhang2019ranksrgan} & 25.244 / 0.692 & 25.243 / 0.692 & 25.304 / 0.694 & 25.297 / 0.694 & 25.306 / 0.694\\
SPSR~\cite{ma2020structure} & 25.200 / 0.708 & 25.201 / 0.708 & 25.263 / 0.710 & 25.252 / 0.710 & 25.258 / 0.710\\
\hline
CrossNet-8$\times$~\cite{zheng2018crossnet} & \textbf{32.914} / 0.933 & 31.509 / 0.911 & 30.610 / 0.893 & 29.928 / 0.878 & 29.441 / 0.865\\
SRNTT-8$\times$~\cite{zhang2019image} & 24.485 / 0.758 & 24.306 / 0.749 & 24.254 / 0.745 & 24.170 / 0.740 & 24.115 / 0.737\\
TTSR-8$\times$~\cite{yang2020learning} & 31.913 / 0.923 & 31.002 / 0.908 & 30.383 / 0.896 & 29.843 / 0.884 & 29.405 / 0.874\\
Cross-MPI & 32.878 / \textbf{0.937} & \textbf{31.736} / \textbf{0.921} & \textbf{30.993} / \textbf{0.909} & \textbf{30.360} / \textbf{0.896} & \textbf{29.852} / \textbf{0.885}\\
\hline
\end{tabular}
\end{center}
\vspace{-6mm}
\end{table*}

\noindent {\bf Reconstruction Loss.} We first define a reconstruction loss to encourage the output $I_{SR}$ to match the ground truth, and we choose L1 per-pixel loss:
\begin{equation}
\mathcal{L}_{rec} = \frac{1}{\beta h\times\beta w\times{c}}\lVert{I_{SR} - I_{GT}}\rVert_1,  \label{con:loss_rec}
\end{equation}
where $(\beta h, \beta w, c)$ is the spatial size of the SR. Moreover, we adopt $L_1$ loss to produce sharper results and easier convergence.

\noindent {\bf Perceptual Loss.} Perceptual loss has been proven to improve visual quality and has been successfully applied to image super-resolution and generation tasks~\cite{dosovitskiy2016generating, johnson2016perceptual, ledig2017photo, zhang2019image, yang2020learning}. Specifically, we adopt the normalized VGG-19~\cite{simonyan2014very} referring to the layer matching from~\cite{chen2017photographic}:
\begin{equation}
\mathcal{L}_{per} = \sum_{l}{\lambda_l\lVert{\phi_l(I_{SR})-\phi_l(I_{GT})}\rVert_1},  \label{con:loss_per}
\end{equation}
where $\{\phi_l\}$ is a set of appointed neural layers of VGG-19 and the weights $\{\lambda_l\}$ are set as the inverse of the number of neurons in each layer.

\noindent {\bf Internal Supervision Loss.} Since a precise correspondence estimation is crucial in our RefSR task and we propose estimating initial alpha map in LR space followed by gradually guided upsampling, it is significant to consider the internal estimation quality. Thus, we propose an internal supervision (IS) loss to monitor the accuracy of the initial estimate of the alpha maps $\mathcal{A}_{init}$. Following the content transfer in Sect.~\ref{section:fusion}, we reshape $PSV_{Ref\downarrow}$ into $\widehat{PSV}_{Ref\downarrow}\in\mathbb{R}^{n\times{c}\times{d}}$ and $\mathcal{A}_{init}$ into $\widehat{\mathcal{A}}_{init}\in\mathbb{R}^{n\times{d}\times{1}}$, where $n = h\times{w}$, and the loss can be expressed as:
\begin{equation}
\mathcal{L}_{is} = \frac{1}{h\times w\times{c}}\lVert{\widehat{PSV}_{Ref\downarrow}\otimes\widehat{\mathcal{A}}_{init}-I_{LR}}\rVert_1, \label{con:loss_is}
\end{equation}
where $\otimes$ denotes batch-wise matrix multiplication. This internal supervision loss encourages the initial alpha maps to produce an accurate content warping at the beginning, which helps our whole pipeline generate more precise SR alpha maps for HR detail transmission.

\subsection{Implementation Details}
\label{section:implementation}

The shared feature extractor contains a residual ASPP (atrous spatial pyramid pooling) block similar to~\cite{wang2019learning} concatenated with a regular residual block. To balance the LR plane-ware attention and guided upsampling procedure, we calculate the plane-ware attention at scale $2\times$ and gradually upsample the initial alpha maps 4 times by our multiscale guided upsampling module to achieve 8$\times$ upsampling. For the multiscale guided upsampling module, we adopt residual blocks and nearest-neighbour interpolation for gradual upsampling, while the guidance is gradually downsampled by residual blocks and convolution with stride=2. For the fusion module, we use cascaded residual blocks to extract HR features of warped $I_{Ref}$ conditioned on $I_{LR\uparrow}$ and merge these features with the upsampled blurry input to produce the final SR result.

During the training, we set the spatial size $h=384, w=768, c=3, d=32, \beta=8$ and use the training set of the RealEstate10K~\cite{zhou2018stereo} dataset. Note that we set $d=32$ to be consistent with \cite{zhou2018stereo,tucker2020single} for MPI generation. The weights for $\mathcal{L}_{rec}, \mathcal{L}_{per}, \mathcal{L}_{is}$ are 1, 1 and 0.2. We train the proposed network in three steps. Specifically, we first pretrain initial alpha maps with internal supervision loss, second the multiscale guided upsampling module is added to warp the reference image, and finally, train the whole network together with all losses. The network is trained with the ADAM solver~\cite{kingma2014adam} for 816.2k, 326k and 472k iterations with learning rates of 0.0002, $\beta_1 = 0.9$, $\beta_2=0.999$ and batch size = 1 on a single TITAN X GPU.

\section{Experiments}
\label{section:exp}
In this section, we introduce two kinds of data from digitally synthesized cross-scale data to optical zoom cross-scale data for quantitative and qualitative comparisons to verify the superior performance of the proposed Cross-MPI. The description of the data is in Sect.~\ref{section:dataset}, and the comparison results of digitally synthesized and optical zoom cross-scale data are presented in Sect.~\ref{section:digital_comparison} and Sect.~\ref{section:optical_comparison}, respectively. The ablation study is then presented in Sect.~\ref{section:ablation}.
\begin{figure*}
\begin{center}
\begin{tabular}{p{3.7cm}<{\centering} p{3.7cm}<{\centering} p{3.7cm}<{\centering} p{3.7cm}<{\centering} }
Ground Truth & RCAN~\cite{zhang2018image} & RankSRGAN~\cite{zhang2019ranksrgan} & CrossNet~\cite{zheng2018crossnet} \\
\hline
Parallax & SRNTT~\cite{zhang2019image} & TTSR~\cite{yang2020learning} & Ours \\
\multicolumn{4}{c}{\includegraphics[width=1.0\linewidth]{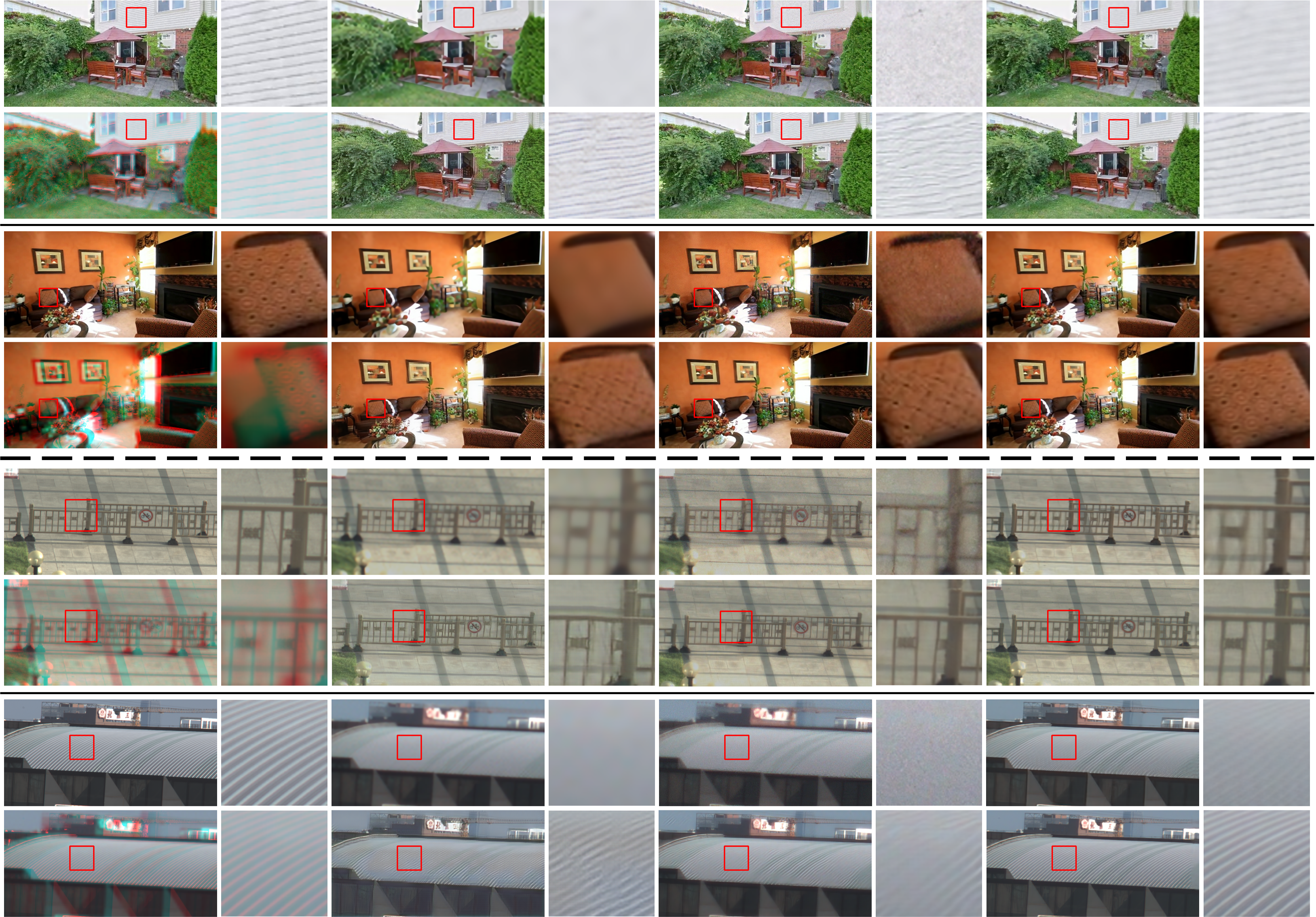}}
\end{tabular}
\end{center}
\vspace{-5mm}
\caption{Visual evaluation (8$\times$) on the RealEstate10K dataset~\cite{zhou2018stereo} (top) and optical zoom-in dataset (bottom).}
\vspace{-6mm}
\label{fig:comparison}
\end{figure*}

\subsection{Datasets}
\label{section:dataset}

The first type of data is the digitally synthesized cross-scale data where the LR images are generated by bicubic downsampling. We use the {\bf RealEstate10K}~\cite{zhou2018stereo} dataset to generate cross-scale data for training (59,637 sequences) and testing (6,546 test sequences). We obtain the LR images through bicubic downsampling (8$\times$). The second type of data is the optical zoom cross-scale data where the HR reference images are generated by optical zoom. Specifically, we collect an {\bf optical zoom-in dataset} with a motorized zoom camera for training and testing and a {\bf cross-scale stereo sequence} collected by a real cross-scale stereo camera setup. In the optical zoom-in dataset, we collected 15 scenes, while each scene contained 30 to 60 content-matched image pairs (with a resolution difference of approximately $10\times$) along with their camera parameters and poses, which were further divided into a training set (13 scenes) and a test set (2 scenes). In the cross-scale stereo dataset, we obtained a set of stereo pairs with focal length ``16 mm-135 mm'' from different views of the scene, introducing not only resolution gaps but also parallaxes between stereo pairs. More details of the camera configurations and calibration procedures can be found in the supplement.

\vspace{-2mm}
\subsection{Comparisons on Digital Synthesized Data}
\label{section:digital_comparison}
\vspace{-1mm}
In this section, we quantitatively compare the performance of Cross-MPI against the latest RefSR~\cite{zheng2018crossnet,zhang2019image,yang2020learning} and SISR~\cite{mdsr,zhang2018image,wang2018esrgan,zhang2019ranksrgan,ma2020structure} methods on the RealEstate10K~\cite{zhou2018stereo} dataset. To make a fair comparison, we re-train CrossNet, SRNTT and TTSR following the procedure according to their papers on the training set of RealEstate10K. Since SRNTT and TTSR do not have 8$\times$ models, we add a $2\times$ subpixel upsampling layer at the end of the main network. For the SISR works~\cite{mdsr,wang2018esrgan,zhang2019ranksrgan,ma2020structure} without 8$\times$ models, we use 4$\times$ models twice and downsample 2$\times$ to achieve 8$\times$ magnification on test images. Moreover, the SR spatial resolution of training set and test set is set to (384,768), consistent with our training procedure.

Table~\ref{table:realestate_num} shows the quantitative comparisons between Cross-MPI against the SISR and RefSR methods. We test cross-scale image pairs for different frame intervals between cross-scale image pairs. Cross-MPI achieves the highest score on the RealEstate10K dataset for 8$\times$ magnification compared with all the tested SISR methods and 
methods are grouped by SISR approaches (top) and RefSR approaches (bottom) under large parallax (frame differences from 11 to 23). When the parallax decreases, the PSNR result is similar to CrossNet, which is suitable for small parallaxes. However, our SSIM is higher because Cross-MPI can preserve the underlying scene structure and generate SR results with better structural performance. Moreover, Cross-MPI is also visually superior to the present SISR and RefSR work, as shown in the upper two cases of Fig.~\ref{fig:comparison}.

\begin{table}
\begin{center}
\caption{Test results on the optical zoom-in dataset.}
\label{table:optical_zoom}
\begin{tabular}{l|c c}
\hline
Algorithm & PSNR / SSIM\\
\hline
CrossNet-8$\times$~\cite{zheng2018crossnet} & 27.830 / 0.798\\
SRNTT-8$\times$~\cite{zhang2019image} & 24.762 / 0.687\\
TTSR-8$\times$~\cite{yang2020learning} & 27.114 / 0.787\\
Cross-MPI & \bf{29.209 / 0.841}\\
\hline
\end{tabular}
\end{center}
\vspace{-4mm}
\end{table}
\begin{figure}[t]
\begin{center}
\includegraphics[width=1.0\linewidth]{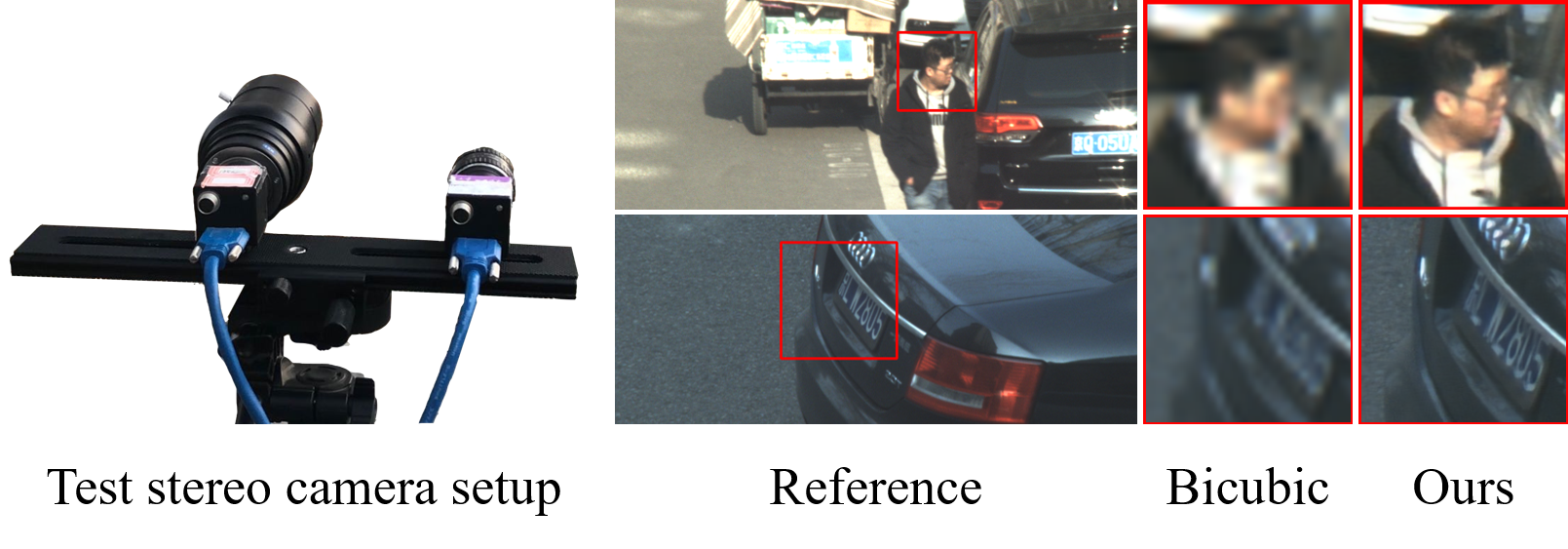}
\end{center}
\vspace{-4mm}
\caption{The real cross-scale stereo results.} 
\label{fig:real_stereo}
\end{figure}

\subsection{Comparisons on Optical Zoom Data}
\label{section:optical_comparison}
\vspace{-1mm}
We also compare our model with the state-of-the-art RefSR models~\cite{zheng2018crossnet,zhang2019image,yang2020learning} on our optical zoom-in dataset. In detail, we fine-tune the pre-trained 8$\times$ models on the training set for 2 epochs and test on the test set. Table~\ref{table:optical_zoom} gives the numerical results, and the last two cases in Fig.~\ref{fig:comparison} give the visual results. Fully utilizing the concealed scene structure, Cross-MPI performs the best numerically and visually on the optical zoom-in dataset, which has a more complicated degradation model than digitally synthesized cross-scale data. To further test the effectiveness of Cross-MPI for real hybrid cross-scale data, we build a cross-scale stereo camera prototype, shown in Fig.~\ref{fig:real_stereo}.

\subsection{Ablation Study}
\label{section:ablation}
\begin{figure}[t]
\begin{center}
\includegraphics[width=1.0\linewidth]{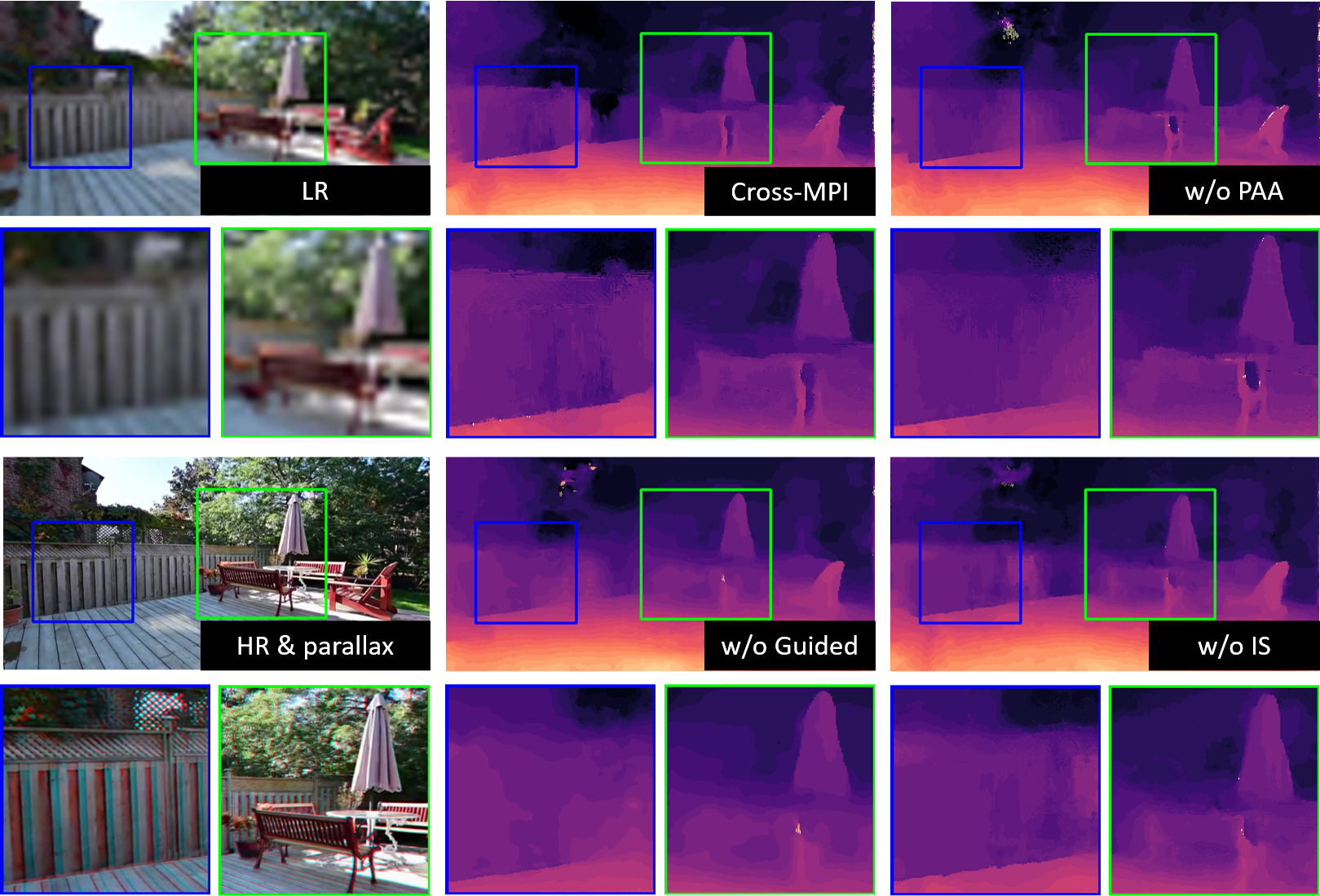}
\end{center}
\vspace{-4mm}
\caption{Qualitative ablation study on Cross-MPI.} 
\label{fig:ablation}
\end{figure}

\vspace{-1mm}
In this section, we verify the effectiveness of different modules in our Cross-MPI network, including plane-aware attention, the multiscale guided upsampling module and the internal supervision loss. We train models with different combinations of these modules. Randomly sampling stereo pairs with frame differences from 3 to 10 as the test set, we calculate the quantitative metrics recorded in Table~\ref{table:ablation} and give an example of the rendered depth in Fig.~\ref{fig:ablation}.

\noindent {\bf Plane-Aware Attention.} To test that the novel plane-aware attention improves the correspondence matching in the whole pipeline, we replace the plane-aware attention with ``concatenation + convolution'' while the rest of the network, as well as the training procedure, remain the same as our full model. As we can see in Table~\ref{table:ablation} and Fig.~\ref{fig:ablation}, the numerical result slightly decreases and some incorrect estimation appears without the proposed plane-aware attention.

\noindent {\bf Multiscale Guided Upsampling Module.} The design of multiscale guided upsampling effectively helps to super-resolve the initial alpha maps. To verify the effectiveness of the guided upsampling, we remove the external guidance path (the lower branch in Fig.~\ref{fig:guided}), and the upsampling procedure becomes purely convolutional upsampling. In Fig.~\ref{fig:ablation}, we can see that the depth becomes blurry without precisely designed guidance.

\noindent {\bf Internal Supervision Loss.} The internal supervision loss helps the network learn a better initial estimation of the alpha maps, which is crucial for the latter SR synthesis and fusion procedure. The numerical and visual results decrease without the IS loss, as can be seen in Table~\ref{table:ablation} and Fig.~\ref{fig:ablation}.

\begin{table}
\begin{center}
\caption{Quantitative ablation study on Cross-MPI.}
\label{table:ablation}
\begin{tabular}{l|c|c|c|c }
\hline
Algorithm & PAA & Guided & IS & PSNR/SSIM \\
\hline
w/o PAA &  & \checkmark & \checkmark & 33.015/0.93266\\
w/o Guided & \checkmark &  & \checkmark & 32.981/0.93269\\
w/o IS & \checkmark & \checkmark &  & 32.982/0.93045
\\
Cross-MPI & \checkmark & \checkmark & \checkmark & \textbf{33.100/0.93275} \\
\hline
\end{tabular}
\end{center}
\vspace{-2mm}
\end{table}
\vspace{-2mm}
\section{Conclusion}
\label{section:con}
\vspace{-1mm}
This paper explores the RefSR problem for real multiscale camera systems. We propose Cross-MPI, an end-to-end network that takes the underlying scene structures as a clue for transferring HR details from one view to the other under a large resolution gap, \eg, $8\times$. The proposed Cross-MPI consists of a plane-aware attention-based MPI module to explicitly and effectively capture correspondences along depth planes, a multiscale guided upsampling module to estimate correspondences in a coarse-to-fine manner with an elaborately designed guided upsampling mechanism, and an SR synthesis and fusion module to adequately transfer HR details and generate the final SR result. Extensive experiments show the superior performance of Cross-MPI over the state-of-the-art in both digital synthesized and optical zoom cross-scale datasets. 

It is worth noting that our method mainly focuses on the SR quality of the matched (non-occluded) regions, while the SR for occluded regions can only rely on the LR input in our design. A possible future work is to solve the occlusion problem by referring to similar textures from the non-occluded areas.

\noindent\textbf{Acknowledgements.} This paper is supported by the National Key Research and Development Program of China [2018YFB2100500] and the NSFC No.61827805 and No.61861166002.

{\small
\bibliographystyle{ieee_fullname}
\bibliography{egbib}
}

\end{document}